\documentclass[pdflatex,sn-mathphys]{sn-jnl}

\jyear{2021}%
\DeclareUnicodeCharacter{2212}{-}
\theoremstyle{thmstyleone}%
\theoremstyle{thmstyletwo}%
\usepackage{natbib}
\theoremstyle{thmstylethree}%

\begin{document}

\title[Advancements in Scientific Controllable Text Generation Methods]{Advancements in Scientific Controllable Text Generation Methods}


\author[1]{Arnav Goel\textsuperscript{*}}
\email{arnav21519@iiitd.ac.in}

\author[1]{Medha Hira\textsuperscript{*}}
\email{medha21265@iiitd.ac.in}

\author[1]{Avinash Anand\textsuperscript{*}}
\email{avinasha@iiitd.ac.in}

\author[1]{Siddhesh Bangar\textsuperscript{*}}
\email{siddheshb008@gmail.com}

\author[2]{Dr Rajiv Ratn Shah}
\email{rajivratn@iiitd.ac.in}

\affil[1]{\textit{Multimodal Digital Media Analysis Lab, IIITD}}

\footnotetext[1]{These authors contributed equally to this work.}





\abstract{The previous work on controllable text generation is organized using a new schema we provide in this study. Seven components make up the schema, and each one is crucial to the creation process. To accomplish controlled generation for scientific literature, we describe the various modulation strategies utilised to modulate each of the seven components. We also offer a theoretical study and qualitative examination of these methods. This insight makes possible new architectures based on combinations of these components. Future research will compare these methods empirically to learn more about their strengths and utility.}




\maketitle

\section{Introduction}\label{sec1} 
The amount of free text on the Internet is enormous many orders of magnitude greater than the number of labelled benchmark data sets. Modern language models (LM) are trained on a huge scale using unsupervised Web data. When generating samples from a language model, we have minimal influence over the resulting text's desired topic, style, sentiment, and other properties. Controlled text creation means producing coherent sentences while maintaining control over various properties. These properties encompass elements such as style, sentiment, formality and intent; demographic aspects like gender or age and the organization of events or information, such as the arrangement of plot summaries. By manipulating multiple text generation properties, diverse objectives can be accomplished. Areas of focus in the field of dialogue response generation include the control of persona, where efforts have been made to manipulate and shape the character or identity expressed in the generated dialogues \citep{zhang2018personalizing, li2016persona}, controlling various response characteristics such as politeness \citep{niu2018polite}, formality, intent \citep{gu2022controllable, jung2022intent}, etc., response grounding in fixed information \citep{zhou2018dataset, dinan2018wizard, ghazvininejad2018knowledge}, and controlling topic sequence \citep{tang2019target, prabhumoye2020love}.

\begin{figure*}[t]
\centering
\includegraphics[width=10cm, height=4cm]{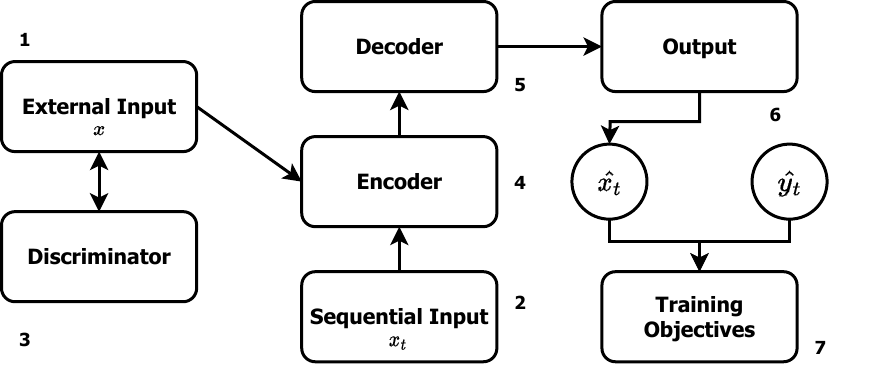}
\caption{Modules Schema for the scientific controllable text generation process.}
\label{schema}
\end{figure*}

Despite the large volume of past research on programmable text production, there is no overarching topic or subject that embraces it all. Each study focuses on specific tasks within particular contexts. Consequently, the challenge lies in determining how to guide a potent unconditioned language model in accordance with personal preferences and desired outcomes. In this study, we provide a novel paradigm that links earlier research and sheds light on many facets of controlled text synthesis. The schema comprises 7 modules that span the pipeline, explain how each part affects the generating process, and explain each technique and method that might provide better insights. We also offer an analysis of the parallels between earlier work that concentrated on particular schema elements we present here.

The proposed schema in this study, illustrated in Figure \ref{schema}, comprises seven modules designed to facilitate the production process. The initial module, labeled as \textbf{(1) External Inputs}, serves as the starting point for the generation process. At each generational time step $t$, the input is sourced from the \textbf{(2) Sequential Inputs} module. Furthermore, this input can be simultaneously shared and trained with the \textbf{(3) Discriminator} module to acquire valuable feedback on the utilized inputs. However, it is important to note that the discriminator should not be employed indefinitely during the development of new architectures.

The \textbf{(4) Encoding Operations} module is responsible for performing consistent operations or computations on each input at every time step. These computations are then propagated forward, generating relevant outputs through the \textbf{(5) Decoding Strategies} module. Subsequently, the \textbf{(6) Output} is projected onto the vocabulary space in order to predict the next token. Finally, the \textbf{(7) Training Objectives} module handles the necessary loss functions for training the generator.

In summary, the proposed schema consists of interconnected modules, each contributing to different aspects of the producing process. The flow begins with external inputs, progresses through sequential inputs, encoding operations, decoding strategies, and output projection, ultimately guided by training objectives.

The contributions of the various modules as well as the techniques and methods for controlling text production, are explained by this schema. This work focuses on using this schema to explain controlled text production, specifically emphasising the application of auto-regressive and uncontrollable language models. Through this effort, new model architectures based on the schema may be designed. This may be accomplished by selecting promising approaches and techniques for each module before fusing them together. This research also provides the linked work in text creation, spotlighting the earlier work while introducing the techniques.

\section{Related Work}
The initial study on related work generation Hoang et al. \cite{hoang2010towards} predates the widespread use of neural networks. Their work focused on summarization, specifically the development of ReWoS (Related Work Summarization). ReWoS employed two distinct strategies for generating summaries: General Content Summarization (GCSumm) and Specific Content Summarization (SCSumm). As a heuristic-based system, ReWoS effectively mapped these strategies to the topic tree structure, which served as the input for the summarization process.

In a subsequent study, Wang et al. \citep{wang2019controllable} explored the explicit extraction of Cited Text Spans (CTS) from cited publications. These CTSs were specific text fragments within the referenced work that were closely related to a particular citation. In addition to utilizing a topic model, the researchers employed a two-layered ensemble model for classifying and extracting the CTS. They employed a greedy algorithm to select candidate phrases, creating connected parts of the text and forming coherent sections. This approach aimed to enhance the controllability and accuracy of the collected and retrieved data.

Addressing the challenge of automatically generating citation texts within academic works, Xing et al. \citep{xing2020automatic} proposed a novel approach. Due to a scarcity of training data, the researchers annotated a dataset and trained an implicit citation extraction algorithm. They suggested the use of a multi-source pointer-generation network with cross-attention mechanisms to effectively tackle this issue. This method allowed for more precise and efficient automatic generation of citation texts.

In contrast to the work by Ge et al. \cite{ge2021baco} and Chen et al. \cite{chen2021capturing}, Luu et al. \cite{luu2020explaining} aims to utilize citation sentences as a form of partial supervision for elucidating the connections between two scientific articles. They explore two approaches in their study. Firstly, they fine-tune the GPT2 model on scientific texts using a specific dataset and employ a neural network to generate concise summaries. Secondly, they establish a direct relationship between the target papers and the articles that both cite them and are cited by them by extracting the citation sentences.

When it comes to constructing the related work section, Chen et al. \cite{chen2021capturing}, building upon the configuration proposed by Hu et al. \cite{hu2014automatic}, utilize the title's keywords to identify relevant themes. They leverage a discriminative feature graph to select pertinent sentences. To overcome the computational complexity of the set cover problem, they introduce a greedy approximation technique that prioritizes phrases containing the most uncovered information at each step. This technique facilitates the efficient construction of the related work section while ensuring the inclusion of critical details from the available sources.

Ge et al. \cite{ge2021baco} extended the approach proposed by Xing et al. \cite{xing2020automatic} by incorporating a graph attention network (GAT) to encode the citation network information. They also utilized a hierarchical bidirectional LSTM to store the citation context and the abstracts of the cited papers.

Beltagy et al. \cite{beltagy2020longformer} introduced SCI BERT, a pre-trained language model specifically designed for scientific text based on BERT. This model was trained on various scientific tasks and datasets from diverse fields. In evaluations, SCI BERT outperformed BERT-Base on several tasks, particularly in computer science, achieving new state-of-the-art (SOTA) results with improvements of +3.55 and +0.49 F1 scores through fine-tuning. The performance of SCI BERT even surpassed the results of previously published BIOBERT \cite{lee2020biobert} on biomedical tasks.

Abura et al. \citep{abura2020automatic} employs a straightforward sequence-to-sequence strategy in their research utilising traditional designs to generate citation sentences from the title and abstract of the referenced work. For this objective, they use Transformer, OpenNMT-py, and the Pointer-Generator Network (PGN). The PGN incorporates copy-attention and coverage methods.

Jadika et al. \citep{jadika2011literature} looked at 20 reviews of the literature obtained from journal articles published in JASIST to establish the fundamental knowledge and comprehension of the referenced papers, source papers, and literature review papers. To grasp and emulate the intricacies of human writing in literature reviews, the researchers meticulously analysed linguistic and content attributes present in reviews extracted from esteemed journals within the field of information science. The study aimed to gain a deeper understanding of the stylistic and substantive aspects employed in the scholarly discourse by subjecting these reviews to a thorough evaluation. This endeavour sought to pave the way for the development of automated systems that can mimic the quality and essence of human-authored literature reviews.

Galactica \citep{taylor2022galactica} is an LLM capable of retaining, integrating, and applying scientific data through its reasoning abilities. This proficiency was developed through extensive training utilizing scientific publications, books, databases, and other resources. Galactica performs better on various scientific problems than the current models. Galactica's corpus is excellent and carefully managed, unlike other language models, which rely on a crawl-based paradigm that is not properly vetted. Without overfitting, they could train on it for several epochs. Upstream and downstream performance increases with the usage of repeated tokens.

Wu et al. \citep{wu2021towards}, Jung et al. \citep{jung2022intent} and Gu et al.  \citep{gu2022controllable} improve upon the limitations in existing citation scientific text generation systems such as those proposed by Xing et al. \citep{xing2020automatic} and Luu et al. \citep{luu2020explaining}. The limitations include not accommodating the ability of authors to summarise multiple studies into one citation sentence and the ability of the authors to control the intent of the citation.

According to Cohan et al. \citep{cohan-etal-2019-structural}, the classification of citation texts involves categorizing them into three intent classes: "Background," "Method," and "Result." In this context, when paper $B$ cites paper $A$, "Background" indicates that $B$ is utilizing $A$'s idea as a foundational element for its own work. "Method" refers to $B$ adopting $A$'s methodology, such as experimentation or dataset preparation. On the other hand, "Result" signifies $B$'s comparison of its findings with those obtained by $A$. These classifications are incorporated in the Sci-Cite dataset, and a structural scaffold is employed to accomplish the categorization.

Gu et al. \citep{gu2022controllable} provides the SciCCG dataset where each citation text is between 5 and 200 words. An intent was randomly chosen from among the three to prepare the baselines. The keywords were chosen using KeyBERT \citep{grootendorst2020keybert} and the relevant sentences using SentenceBERT \citep{reimers2019sentence}. They are extraction techniques based on BERT token embeddings to extract keywords and sentences from a text. This baseline is then compared with the attribute suggestor's keyword and sentence extractor. The latter performs much better due to triple-loss while fine-tuning it. The overall citation generator system gave very high accuracy compared to ground truth citations when all three attributes, along with contextual input, were fed in as input. On human evaluation, the model gave much higher preference to other existing systems and showed appropriate usage of the suggested keywords and sentences. 

\section{External Inputs}
In this part, we go through the many methods that may be used to modify the initialization of the generator to regulate the encoding and decoding process. External Inputs are equivalent to Sequential Inputs in the conventional generating method.

\subsection{Decompose}
It is possible to divide the encoder representation into many subspaces, each representing a property that may be manipulated. The encoder representation is divided into two parts by Liu et al. \citep{liu2018learning}, one of which reflects the document's structure and the other of which contains its semantic content. Balchandran et al. \citep{balachandran2020structsum} utilized this approach to manage structure in abstractive summarization. With regard to the encoder representation's dimensions, this work splits. The technique makes the first n dimensions of the encoder representation for the structure and the last n dimensions to represent meaning. Additionally, Balchandra et al. \citep{balachandran2020structsum} provide quantitative and qualitative analyses of the various document structures that may be learned using this method.

A document is represented by its abstract and title for scientific citation text generation. The encoder representation is divided into subspaces for different permutations with the citing and cited paper. 
Xing et al. \citep{xing2020automatic} decomposes the encoder representation into two different encoders. It uses only the referencing paper's context and the referenced paper's abstract to generate relevant target citations. Both are seen as separate sequences of words encoded by the two different encoders to two separate vectors denoting hidden states. Jung et al. \citep{jung2022intent} encodes the two different documents into the same input text. It varies the presence of tokens representing the abstract and title of the citing paper and thus encodes for the two separately. Gu et al. \citep{gu2022controllable} decomposes the encoder representation into many subspaces, adding subspace to store the local context of the citing paper. Local context refers to the tokens of five sentences before our target citation. Thus, encoder representations can store local context to improve text generation at specific target places.

\subsection{Arithmetic and Linear Transform} 
Concatenating a control vector to the encoder's output is one of the simplest ways to regulate the generation. [Output of the encoder; control vector], where $[a; b]$ implies concatenation, will be the external input of the decoder. Here, the control vector would deliver a potent signal to the generator to direct the generating process.

In Fu et al. \citep{fu2018style}, the encoder creates style-free representations that are solely kept content. The encoder representation is then joined to the style control vector to initialize the decoder. The approach described here is commonly employed to merge information obtained from external sources with the context of a discussion to generate coherent dialogue responses \citep{ghazvininejad2018knowledge, zhou2018dataset, dinan2018wizard}. This method is also used to add controls to consider user preferences. Gu et al. \citep{gu2022controllable} prepends the attribute tokens such as intent, keywords and relevant sentences to encoder input before the local and global context to guide the citation generation process. Wu et al. \citep{wu2021towards} and Jung et al. \citep{jung2022intent} similarly prepend intent tokens and corresponding control codes to ensure intent-controlled citation generation for research papers.

\subsection{Stochastic Changes}
Variational auto-encoder, introduced by Kingma et al. \citep{kingma2013auto}, allows you to extract a continuous latent variable from a Gaussian distribution stochastically. This latent variable serves as the foundation for the initialization of the generator. This idea is used by Bowan et al. \citep{bowman2015generating} to produce sentences from this continuous latent representation. Only the Kullback-Leibler (KL) Divergence training goal may be employed with this method of modifying the encoder state.

\section{Sequential Inputs}
In this part, we go through various methods for influencing the sequential input to the decoder at each time step:

\subsection{Arithmetic and Linear Transform}
Sequential Inputs can undergo the same procedure as Arithmetic and Linear Transforms for External Inputs. By incorporating a few additional control vectors $s$, we can modify the input to the decoder similar to how we alter the initialization. To achieve this, the data at each time step is concatenated with the control vectors. During training, the generator commonly employs the teacher forcing technique \cite{williams1989learning}. At each time step $t$, the predictor determines the word to be generated, denoted as $y$, based on the anticipated word embedding $x$ at $t-1$. Notably, $x$ corresponds to $y-1$. To facilitate this process, the input $x$ is concatenated with the context variable $s$ at each time step $t$, resulting in $\hat{x}= [x; s]$.

\section{Discriminator}
This part discusses the discriminator's function in creating controlled text. Although the discriminator is not always required for training the architecture, its inclusion can offer helpful feedback and enhance the resulting text's content. 

\subsection{External Feedback}
Controlling the external input to the generator is frequently done with a regularizer. A common external feedback approach is employing an adversarial loss to alter the latent space. This effectively controls the encoder's latent space, which is then given to the generator as initialization. A multi-layer perceptron is employed in Fu et al. \citep{fu2018style} to predict the input style labels. Similarly, Wang et al. \citep{wang2019controllable} also uses adversarial loss to regulate the latent representation for style characteristics. To ensure that the meaning representation remains devoid of style indications, the loss function is employed in the study conducted by Romanow et al. \citep{romanov2018adversarial}. It trains a discriminator, which analyses a representation as input and assesses if it produces the specified loss. This technique, like the adversarial loss, employs a motivating loss to verify that the style representation truly includes the stylistic information. The cross-entropy loss is used by Gu et al. \citep{gu2022controllable} to fine-tune SCIBERT \citep{beltagy2020longformer} to steer the process of text production depending on purpose. Based on user-suggested keywords and phrases, triplet loss is utilised to fine-tune and direct the text production process. This use of loss to offer external input ensures that the model generates a citation text that has these features and is more human-friendly than other produced texts.

\section{Encoding Options}
One step in the generator process is the choice of encoding. We will briefly review some of the encoding choices most likely applied to developing controlled text generation.

\subsection{Gradient Based Search}

\paragraph{Auto Prompt \citep{shin2020autoprompt}:}
A novel technique is employed to automate the generation of ideas for various tasks through the use of gradient-based search. This method involves creating a prompt that incorporates task-specific inputs and trigger tokens within a predefined template. These trigger tokens, which are shared among all inputs, possess universal applicability.

To identify the universal trigger tokens, a gradient-guided search approach similar to the one utilized in Wallace et al. \cite{wallace2019universal} is adopted. By optimizing the trigger tokens with respect to the desired output using a universal configuration, all inputs from a given dataset can contribute to the process. Initially, each trigger token's embedding is set to a default value, which is subsequently adjusted to minimize the first-order Taylor expansion of the task-specific loss at the current token embedding.

\paragraph{Prefix Tuning:}
Intelligent prompt design creates effective context, which may result in the desired completion. Li et al. \citep{li2021prefixtuning} proposed the notion of Prefix-Tuning in response to the aforementioned phenomenon. This method includes initialising a limited set of trainable parameters at the start of an input sequence, known as the "prefix." Prefix-Tuning is used to guide a Language Model, allowing it to provide more controlled and focused outputs.

\paragraph{P-Tuning:}
P-Tuning \cite{liu2021gpt} is a technique that involves training continuous prompt embeddings, employing specific alternatives for trainable parameters and architecture. Unlike Prefix-Tuning, P-Tuning only requires the input to function correctly, while facing optimization challenges related to discreteness and association.

\paragraph{Prompt Tuning:}
Prompt tuning \cite{lester2021power} simplifies the concept of prefix tuning by limiting the number of changeable tokens prepended to the input text for each downstream operation to a maximum of $k$. This approach achieves results similar to model fine-tuning, even for large models with billions of parameters. This finding is noteworthy since optimizing and executing large models during inference can be computationally expensive. Prompt tuning proves beneficial for transfer learning when adapting to new domains with learned task-specific parameters, outperforming fine-tuning in addressing domain shift concerns. Additionally, the study demonstrates that prompt ensembling, which involves combining multiple prompts for the same task, further improves performance.

\subsection{Recurrent Neural Networks}
Recurrent neural networks (RNNs) operate sequentially, processing elements one by one while considering previous computations. This iterative nature enables them to maintain a form of memory, facilitating the propagation of contextual information across the network. Although RNNs can theoretically handle arbitrarily long sequences of data, in practice, they often struggle with dependencies beyond a few time steps. To address this limitation, Long Short-Term Memory (LSTM) \citep{hochreiter1997long} units were introduced as a variant of RNNs, incorporating specialized "memory cells" in addition to the standard units. These memory cells enable the retention of information over extended periods, and a set of gates control the input, output, and forgetting of information within the memory. This architectural design allows LSTMs to capture longer-term dependencies effectively. The vanishing gradient problem commonly encountered in RNNs is mitigated by these advancements.

Another variation of RNNs, known as gated recurrent units (GRUs), also employs a similar foundational structure as LSTMs to capture correlations on dynamically adapting time scales. Additionally, GRUs employ a gating mechanism to regulate the flow of information. In contrast to traditional models that utilize multiple memory cells and gates, GRUs achieve comparable functionality with a reduced number of separate memory cells and gates.

Researchers such as Wen et al. \citep{wen2015semantically} have made modifications to LSTMs, introducing methods to control the generation process by incorporating conversation act information. In various text generation tasks, RNNs, LSTMs, and GRUs are commonly employed, as evidenced by the works of Prabhumoye et al. \citep{prabhumoye2020exploring}., Rao et al. \cite{rao2018dear}, See et al. \cite{see2017get}, Zhou et al. \cite{zhou2018dataset}, and Fu et al. \cite{fu2018style}. Despite the utility of these models, their ability to handle long sequences remains a challenge, often necessitating the integration of attention mechanisms to enhance performance on the original sequence.

\subsection{Transformer}
The utilization of the Transformer model, as described by Vaswani et al. \cite{vaswani2017attention}, enables the establishment of global connections between input and output by leveraging attention mechanisms. Within the Transformer architecture, both the encoder and decoder components consist of multiple layers of self-attention and fully connected layers. Specifically, the encoder comprises N identical layers, each containing two sublayers: a self-attention mechanism with multiple heads in the first sublayer, and a position-wise fully connected feed-forward network in the second sublayer. Layer normalization and residual connections are applied to each of these sublayers. Additionally, the decoder includes an additional third sublayer that performs multi-head attention over the output of the encoder stack.

By employing an attention mechanism as its core component, the decoder in the Transformer model can selectively focus on any position within the input sequence. Consequently, computations across the sequence can be parallelized, resulting in improved efficiency. However, it is important to note that the Transformer architecture has not been extensively explored for studying modifications to recurrent neural network (RNN) computing units that incorporate specialized parameters for controlling features such as style, dialogue act, and so on.

\subsection{Pre-trained Language Model} 
Newly pre-trained conditional language models, including XLNet \cite{yang2019xlnet}, GPT \cite{radford2018improving}, GPT-2 \cite{radford2019language}, and GPT-3 \cite{brown2020language}, have been widely utilized in text generation tasks. Researchers have made efforts to enhance these pre-trained models for specific controlled text generation tasks in various studies \cite{sudhakar2019transforming, dinan2018wizard, urbanek2019learning}. However, a recent model called Galactica \cite{taylor2022galactica} has outperformed the more recent GPT-3 model in technical knowledge-related evaluations, achieving a higher score of 68.2\% compared to 49.0\%.

While pre-trained models often generate fluent and grammatically correct content, adapting them for sequence-to-sequence applications like machine translation and abstractive summarization can be challenging. One model that excels in text generation when modified is the denoising autoencoder BART \cite{lewis2019bart}, which utilizes a sequence-to-sequence architecture. On the other hand, T5 \cite{raffel2020exploring} treats every natural language processing (NLP) problem as a "text-to-text" problem, where it takes text as input and produces new text as output, making it well-suited for controlled text generation tasks.

Another approach for controlled language generation is the Plug and Play Language Model (PPLM) introduced by Dathathri et al. \cite{dathathri2019plug}. PPLM combines a pre-trained language model with one or more attribute classifiers, eliminating the need for training from scratch and allowing it to drive text generation based on specific attributes.

Large Language Models (LLMs) are trained on a large corpus of general text. These models have been performing extremely well on text-generation tasks. It gives a promising performance on downstream tasks such as complex reasoning, problem-solving, question-answering, etc. PaLM \citep{chowdhery2022palm} is a 540 billion parameter, dense decoder-only transformer model. This, combined with fine-tuning techniques like Chain-of-Thought Prompting \citep{wei2022chain}, shows remarkable performance on reasoning and understanding-based text generation tasks. LLAMA \citep{touvron2023llama} is a collection of LLMs with parameters ranging from 7 billion to 65 billion. This has been fine-tuned by techniques to create Alpaca \citep{taori2023alpaca}, which is fine-tuned on 52K instruction-following demonstrations generated in the technique called self-instruct \citep{wang2023selfinstruct}. This technique uses a smaller policy Language Model to create similar demonstrations as those in the existing dataset. This creates a larger corpus of data for the bigger language model to be fine-tuned with.

These large language models perform well on text generation tasks as they are trained on a large corpus of texts using attention-based transformers, which helps ensure fluency and diction in the output text. Additionally, these can be fine-tuned using techniques such as prompting to show remarkable improvements on domain-specific downstream tasks. However, these language models often suffer problems such as bias, toxicity and hallucinations which can severely deteriorate accuracy in certain cases.

\section{Decoding Strategies}
During training, these tactics are not employed as a loss aim. Numerous of these goals rely on post-hoc decoding techniques. In this section, we will discuss decoding methods such as Top k-sampling, nucleus sampling, or versions of beam search.
\\
\paragraph{Greedy Search:}
Using a greedy search, which chooses the most likely word at each stage of the sequence of output, is a straightforward approximation. 
Then the decoder's output is projected onto the entire vocabulary space, and we compute the softmax probabilities for each predicted token $y_t$ at time step $t$. We choose the token with the highest softmax probability, which gives us the word with the maximum likelihood or probability at each step $t$ \citep{jung2022intent}. Another technique, as given by \citep{gu2022controllable}, tries to find keywords by ranking candidates' keywords' embeddings based on their cosine similarity with the contextual text embeddings. It uses this ranking to then fine-tune SCI BERT on triplet loss \citep{zhong2020extractive} and chooses the most highly ranked keyword as the output. These are all greedy methods of decoding and choosing outputs, also known as \textbf{greedy decoding}. These methods have the advantage of being very quick, but the final output sequences' quality might not be at its best. 

\paragraph{Beam Search:}
{Greedy search solution might not output the best quality sentence because of only giving locally optimal solutions. Hence, the beam search technique is used. }With a constrained bandwidth, it effectively performs a breadth-first search, one token for each tree level. Beam search extends all descendants of the best candidates at each level of the search tree and maintains a record of the best candidates (referred to as "beam width") at each level. If a beam search encounters the EOS (end-of-sentence) token, it may cease extending a node. However, high-quality generation is not a given with maximization-based decoding.

\paragraph{Top-K Sampling:}
In their work, Fan et al. \cite{fan2018hierarchical} presented a straightforward yet highly effective sampling technique known as Top-K sampling. This approach distributes the probability mass exclusively among the top $K$ words predicted to have the highest likelihood. GPT2 incorporated this sampling strategy, which played a significant role in enhancing its performance in narrative generation.

\begin{figure}[!htb]\centering
   \begin{minipage}{0.48\textwidth}
     \includegraphics[width=.9\linewidth]{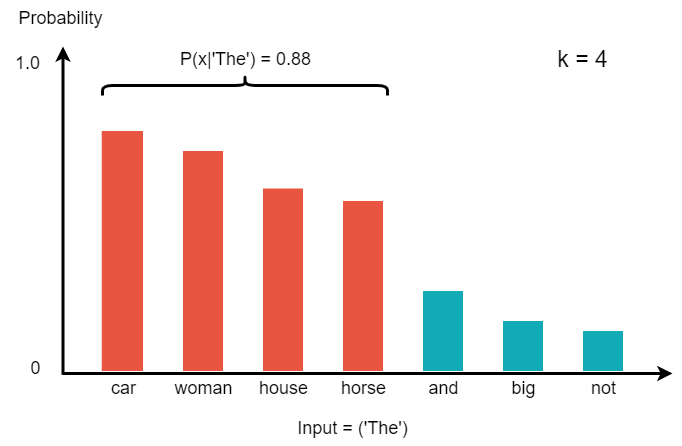}
     \caption{Top-K Sampling Pass 1}\label{Fig:topk1}
   \end{minipage}
   \begin {minipage}{0.48\textwidth}
     \includegraphics[width=.9\linewidth]{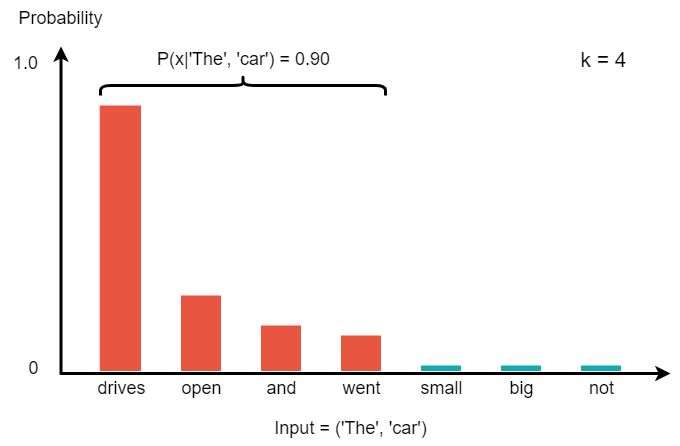}
     \caption{Top-K Sampling Pass 2}\label{Fig:topk2}
   \end{minipage}
\end{figure}

In Figure \ref{Fig:topk1}, we limit our selection pool to 4 words in both sampling phases with $K = 4$. While the first phase defines two-thirds of the entire probability mass, the second step encompasses practically all probability mass. Figure \ref{Fig:topk2} depicts the removal of unnecessary terms such as $'little', 'large', and 'not'$.

\paragraph{Top-P (Nucleus) Sampling:}
Holtzman et al. \cite{holtzman2019curious} introduced the concept of Top-p sampling, which differs from traditional methods by selecting words based on their cumulative probability exceeding a threshold value, denoted as $p$, instead of solely considering the highest probability $K$. This approach involves applying a revised probability distribution to the selected set of words. Consequently, the size of the word set, or the number of words within it, can vary dynamically depending on the probability distribution associated with the subsequent word.

\begin{figure}[!htb]\centering
   \begin{minipage}{0.48\textwidth}
     \includegraphics[width=.9\linewidth]{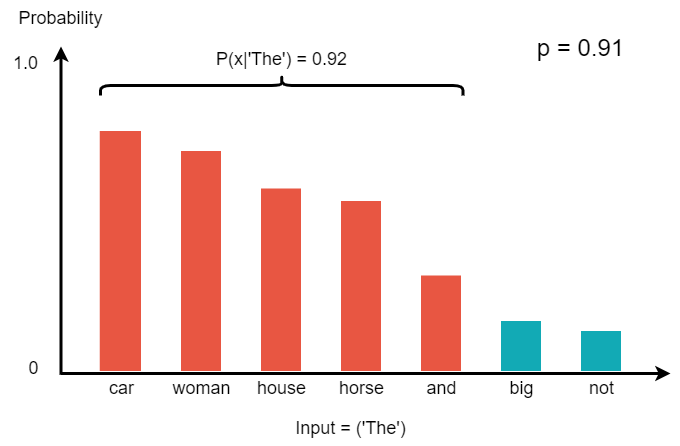}
     \caption{Top-P Sampling Pass 1}\label{Fig:topp1}
   \end{minipage}
   \begin {minipage}{0.48\textwidth}
     \includegraphics[width=.9\linewidth]{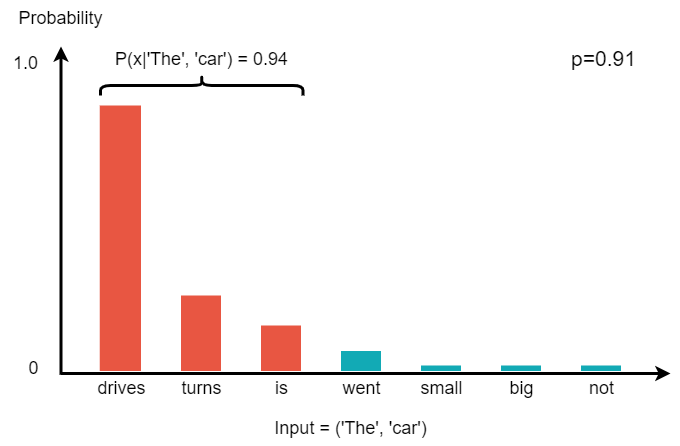}
     \caption{Top-P Sampling Pass 2}\label{Fig:topp2}
   \end{minipage}
\end{figure}

Top-p sampling selects the fewest words that surpass $p=91\%$ when $p=0.91$ is specified. The words are chosen as top-p candidates when the probability mass surpasses the p level. In figure \ref{Fig:topp1}, the approach includes the five most likely terms, but in figure \ref{Fig:topp2}, it only has to select the top three words to surpass 91\%. It can be seen that it maintains a large range of words where the following word is arguably less foreseeable, such as $input = 'The'$, and just a few words where the next word is more predictable, such as $input = 'The', 'car'$.

\paragraph{Penalized Sampling:}
A unique sampling technique was introduced by the CTRL study \cite{keskar2019ctrl} to address the issue of repetitions in generated text. This approach penalizes the scores of previously generated tokens, effectively discouraging the generation of duplicate substrings and mitigating the common failure scenario associated with such repetitions.

\paragraph{Guided Decoding:}
Traditional decoding methods rely on sampling tokens solely based on their likelihood without considering any additional information. However, customizing the candidate ranking score can influence the generated samples based on specific preferences related to subject matter or attitude. By incorporating a selected set of feature discriminators, the ranking score for token selection at each decoding step can be personalized. These discriminators can be designed to evaluate human preferences through heuristics \cite{ghazvininejad2017hafez}, supervised learning \cite{holtzman2018learning}, or real-world testing using reinforcement learning \cite{li2017learning}.

\paragraph{Trainable Decoding:}
Gu et al. \cite{gu2017trainable} proposed a trainable greedy decoding strategy that enables the sampling of sequences from a trained language model in order to optimize a given objective. This method, known as Noisy Parallel Decoding (NPAD), is based on the concept of approximation. To address potential performance degradation, NPAD introduces unstructured noise into the hidden states of the model and performs multiple parallel noisy decodings. Taking this concept further, trainable greedy decoding replaces the unstructured noise with a learnable random variable. A reinforcement learning (RL) agent is employed to predict this random variable using context, previously decoded tokens, and prior hidden states as inputs. In essence, this decoding approach trains an RL actor to manipulate the model's hidden states to achieve desired outcomes.

In a related work, Grover et al. \cite{grover2019bias} trained a binary classifier to distinguish between samples generated by the data distribution and the generative model. By employing a likelihood-free importance weighting (LFIW) technique, this classifier determines the significance weights required for generating a new unnormalized distribution.

\section{Output}
The output of the generator module is projected into the vocabulary space to forecast the next token during the normal generating process. There are numerous techniques of modifying the sequential output before it is projected into the vocabulary space at each time step $t$.

\subsection{Attention}
Attention is a widely employed mechanism that guides the generation process by directing the focus towards the source sequence \cite{chorowski2015attention}. In the context of the generator, the attention module takes the current hidden state and aims to identify relevant source-side information, encapsulated in a context vector, to aid in token prediction. In the case of global attention, the encoder's hidden states are taken into account when computing the context vector. However, it is important to note that this approach incurs a high computational cost, particularly for longer source sequences such as documents.

\subsection{External Feedback}
External feedback can be leveraged to manipulate the latent space of the generator's output. Adversarial loss is one such technique that affects both the output latent space, denoted as $s$, and the external input $x$. Logeswaran et al. \cite{logeswaran2018content} employ an adversarial loss to encourage the generation of words that are both plausible and compatible with desired attributes. The objective of the adversarial loss is to estimate the distribution of sentence and attribute vector pairs $(x, s)$, where the sentence can be either real or intentionally generated.

\section{Training Objectives}
This section explores various methods for regulating the generation process using objective functions. At each generation step, a linear transformation is applied to project the output into the vocabulary space. By applying a softmax function to the transformed output and selecting the token with the highest probability, a token from the vocabulary can be predicted. The predicted token is then compared to the reference token using a loss function. By manipulating the loss function, the generated text can exhibit desired control properties.

\subsection{General Loss}
Cross Entropy Loss: Every text generation mechanism uses this fundamental loss to compare the created tokens to the reference tokens. The generator must anticipate a token from the vocabulary at each time step. Therefore, it might be viewed as a classification issue whereby the number of classes equals the vocabulary size. 

For intent-controlled scientific text generation, the following is used:

$$
L = −\frac{exp(x_{intent}(i_{true}))}{\sum_{i \in intents}{exp (x_{intent}(i))}}
$$
\\
Here $i$ represents the different intents:
\begin{itemize}
    \item $i = 1$ refers to "background" when one paper summarizes the related work and concepts of the other paper.
    \item $i = 2$ refers to "method" when one paper uses a certain method
or dataset of the other paper.
    \item $i = 3$ refers to "result" when one paper compares its results with those of the other paper. 
\end{itemize}
$x_\text{intent}(i)$ is the output obtained when the last hidden state of a prepend token, used to connect the local and global context, is input into the intent prediction header.

\subsection{Prompt Tuning Loss}
Prompting is a new technique for fine-tuning large language models (LLMs). explore into the topic of graph pre-training frameworks, with the goal of smoothly integrating pre-training with downstream tasks for graph neural networks. Formulating a loss function based on a standardised sub-graph similarity template is a major aspect of their technique. This allows for the optimisation of adaptive prompts through the use of task-specific sub-graph representations aided by prompts. The loss function is defined as follows:

$$
\mathcal{L}_{\text{prompt}}(p_t) = -\sum_{(x_i, y_i) \in \tau_t}
\ln(\frac{\exp{(\text{sim}(s_{(t, x_i)}, \tilde{s}_{(t, y_i)})/\tau)}}
{\sum_{c \in Y} \exp{(\text{sim}(s_{(t, x_i)}, \tilde{s}_{(t, c)})/\tau)}})
$$

\begin{itemize}
    \item $\tau_t$ : Labelled Training Set for task $t$. Represented as a set of $(x_i, y_i)$
    \item $x_i$: An Instance or a Node in a Graph
    \item $y_i$: a subset of Y, is the class label for corresponding $x_i$
    \item $\tilde{s}_{(t, c)}$: Representation of a class prototypical subgraph for class c.
\end{itemize}

\subsection{Triplet Loss}
The triplet loss function is used in Gu et al. \citep{gu2022controllable} to fine-tune SciBERT, a language model, with the goal of increasing keyword extraction from text. They hoped to improve scientific text production by recommending appropriate features by utilising the triplet loss. The anchor sample ($x_i$) is matched with a comparable positive sample ($x_p$) and a dissimilar negative sample ($x_n$) in the triplet loss formulation. The goal is to reduce the distance between the anchor and the positive samples while increasing the distance between the anchor and the negative samples:

$$
\mathcal{L}_{\text{triplet}} = \sum_{i=1}^{N} \max\left(d(x_i, x_p) - d(x_i, x_n) + \alpha, 0\right)
$$

$d(a, b)$ represents a distance metric, typically Euclidean distance or cosine similarity. 

$\alpha$ is a margin that defines the desired difference between positive and negative samples.

The phrase within the $max$ function guarantees that the loss is only incurred if the distance between the anchor and positive samples is higher by at least the margin $alpha$ than the distance between the anchor and negative samples. If this criterion is not met, the loss is zero. There are other varieties of triplet loss, such as semi-hard or hard triplet loss, which try to pick negative samples that are more difficult to discriminate against the anchor sample, resulting in greater convergence and discriminative feature embeddings.

The use of triplet loss has found widespread applications when the primary goal is to get a succinct feature space that can successfully discriminate between various entities, such as face recognition and person re-identification.

\subsection{Validation Loss}
Validation loss assesses the model during the validation phase. It measures the difference between expected and actual outputs on a separate validation dataset. Validation loss provides insight into the model's performance beyond the training data by examining how well the model generalises to new, unknown data. It aids in detecting over-fitting and under-fitting by guiding changes in model architecture, hyperparameters, and regularisation strategies. Monitoring validation loss aids in developing models that perform effectively in real-world circumstances.

\subsection{Training Loss}

\paragraph{Unlikelihood Loss}
Including frequent tokens, recurring tokens (or n-grams), and the continuous updating of this set with each token generation helps maintain a pool of negative possibilities \citep{welleck2019neural}. The objective is to minimize repetition across generations, accomplished at both the individual token and sequence levels. This approach can be applied to any task during training alongside the primary objective of maximizing the probability of the target.

\paragraph{KL Divergence Loss}
The Kullback-Leibler (KL) Divergence is a measure of dissimilarity between two probability distributions. It quantifies the difference between distributions $Q$ and $P$, denoted as $KL(P \| Q)$, where the notation $\|$ represents the divergence of $P$ from $Q$. It is important to note that KL Divergence is not symmetric, i.e., $KL(P \| Q) \neq KL(Q \| P)$.

\paragraph{Classifier Loss}
The purpose of this loss is to ensure that the generated tokens possess the desired control attributes. It is important to differentiate this loss, which operates at the token level, from the external feedback loss that operates on latent or hidden representations. The classifier loss employed here is distinct from the external feedback loss utilized in the external input and output modules.

\subsection{Task Specific Loss}

\paragraph{Strategy Loss:}
A conversation strategy-based objective is used by Zhou et al. \citep{zhou2019augmenting} to create replies for negotiation assignments. Ground truth techniques used in this assignment result in more effective negotiations. Given the dialogue history, this loss represents the likelihood that a certain method will be used in the subsequent speech. It directs the generator to match certain techniques with specific answers.

\paragraph{Coverage Loss:}
Text generation systems often face the challenge of producing repeated words or phrases, particularly in tasks involving multi-sentence text generation like abstractive document summarization. To address this issue, See et al. \cite{see2017get} introduced a coverage loss that discourages the model from repeatedly attending to the same locations in the source document.

\paragraph{Structure Loss:}
In the context of abstractive document summarization, Li et al. \cite{li2018improving} proposed two additional loss targets based on sentence-level attention. These objectives, namely structural compression and structural coverage, are specifically designed to improve summarization performance. Structural compression aims to generate a summary sentence by consolidating several distinct source sentences into a concise form. On the other hand, structural coverage focuses on capturing important features from the original text. These loss objectives leverage the structural properties of the document summarization task, evaluating how effectively the generative model can produce shorter and more precise summaries.

\section{Fine Tuning Models}

\paragraph{Conditional Training:}
A conditional language model was trained for 2-step story generation by Fan et al. \citep{fan2018hierarchical}. A narrative writing model then develops a tale based on the sketch the model first generated. A fusion model architecture implements the sketch conditioning method. The story writing model may concentrate on figuring out what the first sketch generation model lacks since the fusion model imposes a type of residual learning. \citep{peng2018towards} experimented with a narrative generator LM that ended valence-conditioned.

\paragraph{RL Fine Tuning:}
It has already been established that fine-tuning sequential models using RL is successful for any arbitrary, non-differentiable reward function \citep{ranzato2015sequence}. There are various drawbacks to the instructor-forcing strategy in sequence generation that may be overcome by RL fine-tuning. During training in instructor forcing, the model minimises the maximum-likelihood loss at each decoding step. During testing, however, the model is expected to create the entire sequence from the beginning. Exposure bias and cumulative mistake can result from the mismatch between training and testing. RL fine-tuning helps overcome these challenges by allowing the model to refine its predictions through reinforcement learning, mitigating the discrepancies between the training and testing phases. For example, BLEU for translation \citep{ranzato2015sequence, wu2016google, nguyen2017reinforcement}; ROUGE for summarization  \citep{ranzato2015sequence, paulus2017deep, wu2018learning}; and a custom measure for story generation may all be directly optimized by RL fine-tuning \citep{tambwekar2018controllable}. 

Li et al. \citep{li2023guiding} tries to use a policy language model (LM) to fine-tune a black-box LLM on downstream tasks using RL. ROUGE scores are used for making the reward function. This generates better keywords and improves performance on summarization tasks. Peng et al. \citep{peng2023check} propose an LLM-Augmenter to enhance language model (LM) performance. Acting as a plug-and-play (PnP) module, it accesses evidence from external databases to improve response generation. Given the impracticality of adjusting the multitude of parameters in large LMs, PnP modules serve as a means to provide automatic feedback and external knowledge, enhancing performance while minimizing computational costs. The suggested approach employs RL to develop a policy function that determines whether to query for evidence, generate a candidate response from the LLM, or deliver the response to the user. This policy augments model-generated responses by comparing them with evidence from the external database. Consequently, the performance of LLMs in addressing queries requiring external databases, such as specific historical questions, is significantly improved.

\paragraph{RL Fine Tuning with Human Preferences:}
In order to ascertain human preferences, the utilization of reward learning is crucial. While quantitative measurement tools such as BLEU or ROUGE are commonly used to compute the overlap of words and n-gram phrases across sequences, they do not necessarily align with the judgments of human evaluators regarding quality. To address this limitation, the approach of reward learning from human input \citep{christiano2017deep} provides a superior method of aligning evaluation metrics with actual priorities. This approach has been successfully applied in various applications, including narrative production \citep{yi2019towards} and summarization \citep{bohm2019better, ziegler2019fine, stiennon2020learning}, where human input is used to train a reward function.

Yi et al. \citep{yi2019towards} collected binary human feedback in four categories for a given dialogue pair (user utterance, system answer), evaluating whether the system response is (1) comprehensive, (2) on topic, (3) interesting, and (4) conducive to continuing the discussion. An evaluator is then trained to anticipate human input and re-rank the beam search samples, thereby improving the model or performing both tasks simultaneously. It is important to note that in their work, supervised fine-tuning rather than reinforcement learning (RL) fine-tuning was employed, and a discriminator loss derived from the evaluator was utilized.

\paragraph{Prompt-Based Fine-Tuning:}
Prompting is a technique that enhances model performance and accuracy on specific downstream tasks without adjusting model weights \citep{weng2023prompt}. Zero-shot prompting involves directly supplying a query to the model and observing its response. In few-shot prompting, the model is provided with $n-1$ query-response examples as a prompt, and in the $n^{th}$ example, only the query is given. This makes the model follow the given query responses as a template and generate the response similarly. Chain-of-Thought (CoT) prompting has shown advancements in LLM performance for complex reasoning tasks \citep{wei2022chain}. CoT prompting aims to generate concise sequences of logical statements that progressively lead to problem-solving. However, preparing human-based annotations for CoT prompting is laborious and challenging.

To address this, Shum et al. \citep{shum2023automatic} propose an Augment-Prune-Select strategy to automate the generation of CoT prompts. This approach generates multiple chains of thoughts and prunes them based on their ability to lead to the correct answer. Zhang et al. \citep{zhang2022automatic} tries to find patterns in mistakes made by the model and clusters them. It thus prevents the frequency of the same type of error demonstration by the model. Chen et al. \citep{chen2023mixture} proposes Mixture of Soft Prompts (MSP). In-context learning is used to prompt a large black-box language model. Thus, they use the model as a data augmentation tool rather than directly predicting the answer. The exemplars generated from the LLM train a smaller policy LM to generate the final answers. Taori et al. \& Li et al. \citep{taori2023alpaca, li2023guiding} use a policy LM to fine-tune a larger language model by prompting techniques. Using LMs or LLMs to fine-tune LLMs is a well-known technique which has improved their accuracy on various downstream tasks and helped automate the process of creating large datasets.

\paragraph{Un\-likelihood Training:}
During language model training, the maximization of log-likelihood loss can lead to an imbalanced token distribution, which cannot be adequately rectified through decoding techniques alone. Specifically, when deterministic decoding is employed, these models tend to generate high-frequency words while neglecting low-frequency terms excessively. In essence, they exhibit overconfidence in their predictions. The training objective is explicitly modified by incorporating preferences for less desirable content through probability training to address this issue. This approach aims to mitigate the bias towards high-frequency words and promote more balanced and accurate generation \citep{welleck2019non}.

\section{Evaluation}
An essential phase in the automated summarizing process is evaluating the resulting text. In this part, we examine how the suggested systems operate and how they are assessed in relation to their starting points.

\paragraph{ROUGE:}
ROUGE is a widely adopted evaluation method used for assessing the quality of automated summarization. It quantifies the level of overlap between candidate summaries and reference summaries authored by humans. The ROUGE measures encompass various statistics, including ROUGE-L, ROUGE-W, ROUGE-N (e.g., ROUGE-1, ROUGE-2), and ROUGES \citep{lin2004rouge}.

\paragraph{BLEU:}
BLEU is an algorithm commonly employed for evaluating the accuracy of machine-translated content across different natural languages. Typically, individual translated segments or sentences are scored by comparing them against a set of accurate reference translations. The resulting scores are then averaged to provide an approximation of the overall translation quality. It is important to note that BLEU evaluation does not consider grammar or higher-level semantic aspects. The output of BLEU is a value between 0 and 1, where higher scores indicate greater similarity between the candidate text and the reference texts \citep{papineni2002bleu}.

\paragraph{F1 Scores}
The F1 score is a widely used evaluation metric in machine learning for quantifying the accuracy of a model's predictions on a given dataset. It combines precision and recall scores to provide a comprehensive assessment of the model's performance.

In the context of evaluating generated responses in the Wiki-QA dataset, the model's outputs are evaluated using token-level precision, recall, and F1 scores, which are calculated by comparing them to annotated answers.

\paragraph{BLEURT Scores}
The BLEURT Score is another evaluation metric employed in assessing Natural Language Generation tasks. It takes a pair of sentences as input and produces a score that indicates the fluency of the generated text with respect to a reference text. It also assesses whether the generated text can effectively convey the same meaning as the reference text. The BLEURT Score has demonstrated state-of-the-art agreement with human judgments on machine translation benchmarks.

\paragraph{SciBERT Score}
The SciBERT Score is designed specifically for evaluating the quality of generated text in the scientific domain, considering its relevance and coherence within that domain. This evaluation metric leverages the domain-specific knowledge captured by the SciBERT language model \cite{beltagy2020longformer} by comparing the generated text to a reference text or a set of annotated reference texts. The SciBERT Score serves as a statistical measure for evaluating the effectiveness of scientific and citation text generation methods.

By providing an informative assessment of the generated content's quality and its alignment with the scientific domain, the SciBERT Score enables researchers and developers to compare different models and drive advancements in text generation systems tailored specifically for scientific applications.

\paragraph{METEOR}
The evaluation of translation quality using the Metric for Evaluation of Translation with Explicit Ordering (METEOR) involves the calculation of similarity between a machine-generated translation and a reference translation. This similarity assessment is based on comparing n-grams, which are consecutive sequences of n words or tokens within a text. N-grams have proven valuable in various applications, including text generation, text analysis, and sentiment analysis.

When evaluating machine-generated answers, the METEOR metric functions by computing a score that considers the matching of words between the generated output and a provided reference. In cases where multiple references are available, the generated output is independently scored against each reference. Consequently, the pair with the highest score is selected as the most suitable match.

\paragraph{Additional Metrics}
\begin{itemize}
    \item Shum et al. \citep{shum2023automatic} uses \textbf{Exact Match Accuracy}. This is done after removing special characters and symbols and checking if the generated citation text has the exact tokens as the ground truth.
    \item Other measures often used to measure the model performance include $sentence-BLEU$, $BERTscore$ and $COMET$.
\end{itemize}

\section{Future Work and Recommendations}
Although transformer-based language models have significantly improved scientific text generation and other downstream tasks, they continue encountering issues like hallucinations, bias, and inaccuracy. Furthermore, most language models trained on broad web-based corpora (LLMs) struggle with domain-specific tasks and cannot incorporate substantial user input for controllability. Here, we comment on future advancements in controllable scientific text generation and present our recommendations based on the findings of our survey:

\paragraph{Controllable Text Generation:}
Gu et al. \citep{gu2022controllable} presents a comprehensive approach to multi-system controllable text generation. They propose identifying specific attributes that users can control to guide the text generation process according to their requirements. These attributes include intent, keywords, and relevant sentences. To enhance the diversity of choices, we suggest fine-grained divisions of intent. For example, the "method" intent can be subdivided into using the dataset and the model described in the cited paper. Moreover, instead of relying on relevant sentences for context, we propose using the specific passage from the referenced paper associated with the intent to generate target citations that provide contextual information. Fine-tuning techniques like Reinforcement Learning from Human Feedback (RLHF) will play an important role here. Currently, proposed systems cannot generate a citation sentence citing multiple documents \citep{wu2021towards}. This can be added as a feature that the author can control and influence through a prompt-based system.
\paragraph{Language Models:}
Large Language Models (LLMs) can perform with improved accuracy over pre-existing systems on domain-specific tasks such as scientific text generation by fine-tuning with those curated corpora of data sets. With the advent of various open-source models with varying numbers of parameters, future work will involve their quantisation to improve access and democratisation. Newer models such as GPT-4, Llama, Alpaca, PaLM \citep{openai2023gpt4, touvron2023llama, taori2023alpaca, chowdhery2022palm} and models specifically fine-tuned on scientific texts should be given higher preference for designing systems upon. Model architectures can be re-looked by opting for different encoding and decoding strategies. Newer strategies not solely based on greedy methods promise to make these processes more efficient and accurate.

\paragraph{Prompting: }
As mentioned earlier, prompting is a valuable technique for fine-tuning large language models to enhance performance without requiring weight updates. One such technique, Chain of Thought Prompting \citep{wei2022chain} along with other discussed methods, helps address hallucinations and improves the model's reasoning abilities. Recent studies by Long et al. \& Zhou et al. \citep{long2023large, zhou2022least} introduce novel prompting techniques, namely Least-to-Most Prompting and Tree of Thoughts Prompting, respectively. These techniques build upon the limitations of Chain of Thought Prompting and further improve model performance. By treating text generation as a reasoning task, these techniques hold promise in enhancing the controllability of language models. In a chain-of-thoughts-like manner, keywords and relevant sentences can be incorporated to guide the model towards generating citation text that closely resembles the ground truth.

\paragraph{Retrieval: }
Lazaridou et al. \citep{lazaridou2022internetaugmented} looks at document retrieval to augment LLMs. It proposes using web search for the same. This has been applied in closed-book question-answering previously. Training LLMs on a specific and fixed corpus of data can be challenging. Alternatively, employing retrieval-augmentation methods, as examined by Soong et al. \citep{soong2023improving}, within a specific domain can simplify the task. These methods retrieve relevant context from domain-specific corpora based on user queries. Subsequently, this extracted information is used as context to seed the LLM, constraining the model's responses to the retrieved text. This innovative technique shows promise in improving model performance and reducing hallucinations by limiting the model's domain space without incurring the time and resource expenses associated with training LLMs.

\section{Conclusion} 
The previous work on controllable text generation is organised using a new schema we provide in this study. Seven components make up the schema, and each one is crucial to the creation process. To accomplish controlled generation for scientific literature, we describe the various modulation strategies utilised to modulate each of the seven components. We also offer a theoretical study and qualitative examination of these methods. This insight makes new architectures based on combinations of these components possible. Future research will compare these methods empirically to learn more about their strengths and utility.

\bibliography{sn-bibliography}



\end{document}